\documentclass[pdflatex,sn-mathphys-num]{sn-jnl}

\usepackage{graphicx}%
\usepackage{multirow}%
\usepackage{amsmath,amssymb,amsfonts}%
\usepackage{amsthm}%
\usepackage{mathrsfs}%
\usepackage[title]{appendix}%
\usepackage{xcolor}%
\usepackage{textcomp}%
\usepackage{manyfoot}%
\usepackage{booktabs}%
\usepackage{algorithm}%
\usepackage{algorithmicx}%
\usepackage{algpseudocode}%
\usepackage{listings}%

\theoremstyle{thmstyleone}%
\theoremstyle{thmstyletwo}%

\theoremstyle{thmstylethree}%
\usepackage{makecell}
\begin{document}

\title[Large Language Models Assisting Screening of Pediatric Depression]{LLM Assistance for Pediatric Depression}


\author[2]{\fnm{Mariia} \sur{Ignashina}}\email{m.ignashina@qmul.ac.uk}

\author[1]{\fnm{Paulina} \sur{Bondaronek}}\email{p.bondaronek@ucl.ac.uk}

\author[3]{\fnm{Dan} \sur{Santel}}\email{daniel.santel@cchmc.org}

\author[6]{\fnm{John} \sur{Pestian}}\email{john.pestian@cchmc.org}

\author*[1]{\fnm{Julia} \sur{Ive}}\email{j.ive@ucl.ac.uk}

\affil[1]{University College London, Institute of Health Informatics, UK}
\affil[2]{Queen Mary University of London, School of Electronic Engineering and Computer Science, London, UK}
\affil[3]{Division of Biomedical Informatics, Cincinnati Children's Hospital Medical Center, University of Cincinnati, Cincinnati, OH, USA}
\affil[6]{Division of Biomedical Informatics, Cincinnati Children's Hospital Medical Center, University of Cincinnati, Cincinnati, OH, USA}


\abstract{\textbf{Introduction:} 

Depression is a complex mental health condition, particularly prevalent among young people aged 10–24, a group experiencing a sharp rise in cases.  Traditional screening methods, such as the PHQ-9, are particularly challenging for children in pediatric primary care due to practical limitations. AI has potential to help but the scarcity of annotated datasets in mental health, combined with the computational costs of training, highlights the need for efficient, zero-shot approaches. Large Language Models (LLMs) offer a promising, computationally affordable zero-shot solution by extracting relevant text segments from electronic patient notes to support clinicians in identifying depressive symptoms, as well as for downstream AI. In this work, we investigate the feasibility of state-of-the-art LLMs for depressive symptom extraction in pediatric settings (ages 6–24). This approach aims to complement traditional screening and minimize diagnostic errors. 

\textbf{Methods:} 
We examined free text of the EHRs of pediatric patients with the diagnosis of depression or related mood disorders (age groups 6-24, 1.8K patients) from at the Cincinnati Children's Hospital Medical Center. We noticed drastic inconsistencies in the application and documentation of PHQ-9 screening highlighting the difficulty in obtaining comprehensive diagnostic data in these conditions. We manually annotated notes for 22 patients with 16 depression-related symptom categories. We leveraged the combination of Beck's Depression Inventory (BDI) and the PHQ-9 to develop tailored categories specifically suited for pediatric depression symptoms. We then applied three state-of-the-art Large Language Models (LLMs) (FLAN T5, Llama and Phi) to automate the identification of these symptom categories.

\textbf{Results:}
Our findings show that all LLMs are 60\% more efficient than word match with Flan leading in precision (average F1: 0.65, precision: 0.78), excelling in the extraction of more rare symptoms like ``sleep problems'' (F1 0.92) and ``self-loathing'' (F1 0.8). Phi strikes a balance between precision (0.44) and recall (0.60), performing well in categories like ``Feeling depressed'' (0.69) and ``Weight change'' (0.78). Llama 3 with the highest recall (0.90), overgeneralizes symptoms, making it less suitable for this type of analysis. Challenges include the complexity of clinical notes and overgeneralization from PHQ-9 scores. The main challenges faced by LLMs include navigating the complex structure of clinical notes with content from different times in the patient trajectory, as well as misinterpreting elevated PHQ-9 scores. We finally demonstrate the utility of symptom annotations provided by Flan as features in an ML algorithm which differentiates depression cases from controls with high precision of 0.78 with major performance boost as compared to a baseline not using these features. 

\textbf{Discussion:}
This study highlights the strengths of LLMs in addressing data scarcity and heterogeneity by extracting depressive symptoms from free text of  pediatric EHRs with high precision. The computational efficiency of FLAN-T5 further supports its potential for deployment in resource-limited clinical settings. Focus on PHQ-9 and BDI symptoms limits the scope of detected depressive symptoms. Additionally, our findings are constrained to the pediatric age group, requiring further validation for broader populations and other mental health conditions. Despite these limitations, the study demonstrates the potential of LLMs to enhance depression screening, improve diagnostic consistency, and provide a transparent, interpretable tool for mental health clinicians.
}




\maketitle

\section{Introduction}\label{intro}

\paragraph{}
Depression is a complex mental health condition that affects 4,7\% \cite{Herrman2022-pm} across various age groups, including children, adolescents, and young adults. It is typically characterized by persistent sadness, loss of interest or pleasure (anhedonia), and a broad range of emotional, cognitive, and physical symptoms. Unlike temporary emotional fluctuations, depressive disorders are marked by their severity, functional impairment, and resistance to relief from everyday positive experiences or social interactions \cite{Zalsman2006-ps}. 

Young people (ages 10–24) are particularly vulnerable to depression, as it coincides with critical stages of social, emotional, and cognitive development. In recent years, there has been a sharp rise in depression within this age group, especially among females. Many young individuals also exhibit ``subthreshold depression''\cite{Rodriguez2012-dl}, where depressive symptoms are present but do not meet the full diagnostic criteria for major depressive disorder (MDD).

A commonly used tool for assessing depression is The Patient Health Questionnaire (PHQ-9). The PHQ-9, developed by \citet{Kroenke2001-se}, is a validated screening instrument based on the Diagnostic and Statistical Manual of Mental Disorders (DSM-IV) criteria for major depressive disorder. It includes nine questions where individuals rate the frequency of depressive symptoms over the past two weeks on a scale of 0 to 3. The total score indicates the severity of depression, with higher scores reflecting greater impairment.

However, screening is difficult to implement in the pediatric primary care setting, with traditional approaches that rely on multiple informants often producing discrepant findings \cite{Behrens2019}. Beyond this, clinicians in the primary care setting vary in how they consider risk factors and interpret symptoms \cite{Tulisiak2017}, and clinicians may struggle to disentangle overlapping symptoms \cite{Strawn2021}.

Most research on detecting depression using machine learning focuses on analyzing social media data~\cite{Liu2022-jq}, where challenges like metaphorical language, sarcasm, and data scarcity make the task particularly complex. Inherent data sparsity problem had led to semi-automatic approaches to data collection via identifying self-expressions of mental illness diagnoses~\cite{coppersmith-etal-2014-quantifying}.

Early studies, such as~\citet{Yazdavar2017-ih}, address the depression detection problem by leveraging vocabulary matches and traditional models like Latent Dirichlet Allocation (LDA) to identify depression-related topics based on PHQ-9 criteria. Building on this foundation,~\cite{Yadav2020-sc} explores the complexities of figurative language, such as metaphors and sarcasm, in depressive social media posts using fine-tuned Transformer-based BERT models to handle this language. More recent studies such as~\cite{mendes-caseli-2024-identifying} emphasise that the detection of subtle, figurative expressions of depression remains a significant hurdle even for the powerful pre-trained models.

Large Language Models (LLMs) have demonstrated significant potential in mental health care due to their advanced linguistic capabilities and generalization ability~\cite{Hua2024}. A key strength of LLMs is their capacity for zero-shot analysis, where they can process and extract insights without requiring annotated training data. This capability is especially valuable in mental health, where annotated datasets are limited due to ethical and privacy concerns. In healthcare applications, smaller models like FLAN-T5 have shown promising results in zero-shot evidence extraction, identifying cases such as postpartum hemorrhage more effectively than traditional methods~\cite{Alsentzer2023}. Similarly, in social science research, FLAN-T5 has outperformed human crowdworkers in producing high-quality free-form explanations, highlighting its potential for zero-shot data annotation~\cite{ziems-css}. 

Despite these promising capabilities, significant challenges remain when using LLMs directly in mental health contexts. Key concerns include hallucinated content, biased outputs, and random guesses when models attempt to classify data without adequate context~\cite{mesko-topol-23}. These issues undermine their reliability as standalone diagnostic tools, even when human oversight is involved. To address these limitations, a safer approach is to leverage LLMs for evidence extraction rather than classification. By focusing on extracting text-based insights for clinical interpretation, this method reduces the risks associated with hallucinations and inaccuracies while preserving the benefits of automated data processing. 

In general domains, LLMs have proven effective for retrieving supporting evidence in question-answering tasks~\cite{Huo2023}. These models excel at extracting relevant information from large text corpora, which can be further analyzed by professionals to support clinical decision-making. This application of LLMs as linguistic search tools presents a practical and low-risk use case in healthcare settings.  

Recent studies have applied LLMs to annotate symptoms of mental health conditions, such as depression, from online posts and psychological interviews~\cite{chim-etal-2024-overview,wang-etal-2024-explainable,Lan2024, Rosenman2024}. These symptom annotations can serve as valuable inputs for downstream analytical models, enhancing the overall robustness of mental health screening. To optimize computational costs, pre-filtering techniques are commonly employed before applying LLMs, ensuring efficient and cost-effective processing. 

AI, particularly LLMs, has the potential to significantly enhance mental health screening procedures, especially given the challenges posed by the lack of comprehensive data and the nuanced variability in depressive symptom presentations. LLMs, with their advanced linguistic capabilities, become essential in this context, as they can efficiently extract and interpret relevant information from unstructured text.

\paragraph{Study Aim and Objectives }

In this work, we focus on assessing the feasibility of using state-of-the-art large language models (LLMs) to assist clinicians in identifying depressive symptoms from free-form text in pediatric electronic health records (EHRs). Rather than relying on LLMs for diagnostic decisions, the models are used to extract potentially relevant text segments, ensuring that clinicians maintain full control over interpretation and decision-making. This approach adheres to ethical principles by preserving human judgment in the diagnostic process, offering a supportive tool for mental health screening in the challenging pediatric setting (ages 6-24).  

This study aimed to evaluate the feasibility of using state-of-the-art LLMs to support mental health diagnostics in pediatric primary care by extracting relevant text segments from EHRs for depressive symptom detection based on PHQ-9 criteria. 

The key objectives were to: 1) Assess the feasibility of zero-shot LLMs in identifying depressive symptoms in free-text clinical notes, with a specific focus on pediatric populations, 2) Compare and benchmark the zero-shot performance of three leading LLM models in extracting PHQ-9 symptom-related information, using manually labeled EHR records as ground truth, 3) Demonstrate the potential clinical value of LLM-driven evidence extraction to improve mental health screening and reduce data sparsity issues using an example of an interpretable AI-based depression screening tool. 

\section{Methods}

\paragraph{Ethical Considerations }

This study was approved by the Institutional Review Board of Cincinnati Children's Hospital as STUDY2020-0942.

\paragraph{Study Design }

This study involved a retrospective cohort analysis of pediatric electronic health records (EHRs) collected from a major pediatric institution over a 10-year period. We apply a range of state-of-the-art NLP methods to these data. 

\paragraph{Data Collection }

We used a foundational database comprising electronic health record (EHR) data from the Cincinnati Children’s Hospital Medical Center (CCHMC) Epic Link, an online platform that provides real-time access to patient charts and clinical results. This database includes approximately 1.3 million unique pediatric patients treated at CCHMC between January 1, 2009, and March 31, 2022, with a total of 63 million clinical notes. 

For this study, we focused on clinical notes recorded between January 2011 and December 2021. PHQ-9 scores were identified by searching for the pattern ``PHQ-9 Total Score:'' within the notes, resulting in 2\% of the notes containing such scores. 

\paragraph{Inclusion criteria:}
\begin{enumerate}
    \item Patients who had a depression diagnosis with at least one visit recorded in the EHR during the 18 months prior to diagnosis;
    \item Diagnoses were identified using ICD-10 codes corresponding to depressive disorders and related mood disorders, including major depressive disorder and its subtypes~\footnote{F32.0, F32.1, F32.2, F32.3, F32.9, F33.0, F33.1, F33.2, F33.3, F33.8, F33.9, F32.81, F32.89, F32.5, F33.40, F33.41, F33.42, F32.4; or these ICD9 codes: 296.24, 296.2, 296.2, 296.25, 296.21, 296.3, 296.22, 296.34, 296.36, 296.31, 296.3, 296.32, 296.26, 296.35, 296.23, 296.33};
    \item All clinical notes attached to visits prior to the diagnosis date were included for analysis.
\end{enumerate}

\paragraph{Exclusion criteria:}

\begin{enumerate}
    \item Patients with incomplete visit records or missing key clinical data were excluded from the analysis;
\end{enumerate}

\paragraph{Cohort Characteristics }

The cohort included 28,172 pediatric patients, aged between 6 and 24 years. The smallest representation is in the 24-year-old group, while the largest is in the 15-year-old group. PHQ-9 completion was more prevalent among older patients, particularly those aged 16 and above, with an average completion rate of 55\% across all age groups (at least one full PHQ-9 questionnaire completed). PHQ-9 documentation beyond the total score was inconsistent, with a notable 67\% drop in documentation for the 15-year-old group. Additionally, proper PHQ documentation was missing for approximately 2\% of patients. See detailed statistics, including age-specific PHQ completion rates (see Table~\ref{tab:overall_phq_stats}).

On average, approximately equal proportions of patients (55\%) completed at least one PHQ-9 or one PHQ-2, with PHQ-9 completion becoming more prevalent in older age groups. Among patients aged 16 and above, 70\% completed at least one PHQ-9, compared to 46\% who completed at least one PHQ-2. 

However, there is a noticeable drop in completion rates for more than one PHQ, with an average decline of 56\% between those who completed one and those who completed two or more PHQs. The average drop in completion rates from answering the first two questions of at least one PHQ to completing at least two PHQs was 38\%, with the largest observed drop being 69\%, underscoring the difficulty in achieving full diagnostic data. A similar decline was observed for full PHQ-9 completions, with an average drop of 35\%. This drop highlights the challenges clinicians face in gathering comprehensive depression-related data. 

\paragraph{Factors Contributing to PHQ-9 Data Sparsity }

Several factors contributed to the observed data sparsity. PHQ-9 tests were commonly administered only in specific departments, such as Primary Care, limiting their broader use. Additionally, phone call scripts often provided supplementary insights into the onset of depression that were not captured through PHQ-9 questionnaires. Younger patients frequently found the questionnaire confusing, leading to incomplete or incorrectly answered items. 

Certain symptoms assessed by the PHQ-9, such as being fidgety or restless, were particularly difficult for individuals to self-evaluate. External motivations, such as fear of stigmatization, could lead to inaccurate responses. In some instances, patients left the hospital before completing all required procedures. Furthermore, PHQ-2 screening protocol dictated that patients testing negative for depression symptoms in the first two questions did not proceed to the remaining seven questions of the PHQ-9, further contributing to the incomplete data. 

\begin{table}
\begin{center}
\scalebox{0.9}{
\begin{tabular}{p{5mm}|p{10mm}|p{10mm}|p{10mm}|p{10mm}|p{10mm}|p{10mm}|p{10mm}|p{10mm}|p{10mm}}
Age Bin & \# patients in cohort & \# visits with PHQ & \# patients with recorded PHQ & \% with at least one PHQ & \% with at least one PHQ-9 & \% with at least one PHQ-2 & \% with at least two PHQ & \% with at least two PHQ-9 & \% with at least two PHQ-2 \\\hline
6 & 94 & 13 & 3 & \bf 100\% & 8\% & \bf 92\%  & 23\% & 0\% & 23\% \\
12 & 2340 & 642 & 507 & 98\% & 17\% & 83\%  & 45\% & 0\% & \bf 45\% \\
13 & 3323 & 937 & 1076 & 99\% & 46\% & 65\% & 35\% & 17\% & 11\%\\
14 & 4015 & \bf 973 & 1274 & 98\% & 52\% & 56\%  & 36\% & 17\% & 8\% \\
15 & \bf 4448 & 931 & 1491 & 98\% & 56\% & 53\% & 36\% & 20\% & 11\% \\
16 & 4296 & 783 & \bf 1565 & 98\% & 70\% & 46\% & 52\% & \bf 35\% & 12\% \\
17 & 3463 & 596 & 1452 & 96\% & 70\% & 48\% & 55\% & \bf 35\% & 15\%\\
18 & 1201 & 285 & 1010 & 98\% & 76\% & 51\% & 60\% & \bf 35\% & 16\% \\
19 & 677 & 165 & 685 & 98\% & \bf 81\% & 49\% & \bf 61\% & \bf 35\% & 22\% \\ 
20 & 454 & 119 & 465 & 99\% & 78\% & 46\% & 59\% & 30\% & 22\%\\ 
21 & 255 & 62 & 193 & \bf 100\% & 69\% & 55\% & 50\% & 26\% & 15\% \\ 
22 & 147 & 27 & 83 & \bf 100\% & 52\% & 63\% & 30\% & 11\% & 11\%\\ 
23 & 107 & 13 & 33 & \bf 100\% & 69\% & 31\% & 8\% & 0\% & 8\%\\ 
24 & 70 & 8 & 19 & \bf 100\% & 62\% & 38\% & 12\% & 12\% & 0\%\\
\hline
Av. & 1778 & 397 & 704 & 98\% & 55\% & 54\% & 42\% & 20\% & 16\%\\
\end{tabular}}
\caption{\label{tab:overall_phq_stats} Overall statistics over the extracted PHQ data. \# denotes number. \%, with at least one PHQ illustrates that some of the PHQ tests that were taken, were not recorded properly and for about 2\% of patients the data is missing. PHQ-9 denotes questionnaires with all the nine questions filled-in. PHQ-2 denotes questionnaires with the first two questions filled-in.}
\end{center}
\end{table}

\subsection{Symptom Annotation Process }

The annotation process was primarily guided by the 21 questions from Beck’s Depression Inventory (BDI)~\cite{kovacs1977empirical}, which were further augmented with additional symptoms from PHQ-9. Domain experts considered the BDI’s comprehensive scope better suited for identifying depression in pediatric patients. Additionally, PHQ-9’s semantics posed interpretability challenges for large language models (LLMs), reinforcing the need to use BDI as the primary framework. 

A total of 85 clinical notes from Cincinnati Children’s Hospital Medical Center were manually annotated. These notes were sourced from departments including Adolescent Medicine, Endocrinology, Primary Care, Pediatrics, and Gastroenterology, covering 22 patients (13 females and 9 males) aged 15-17 years. The average note length was 425 words, and the dataset maintained a balanced class distribution, with 50\% of the notes containing at least one symptom. 

Annotations were performed at the sentence level, explicitly marking the presence or absence of depressive symptoms. This process resulted in binary labels (positive or negative) for each symptom across the entire note. For example, a note was labeled as positive for ``Irritability'' if any sentence within the note contained evidence of irritability. 

\subsection{Challenges in Annotation }

Several challenges were encountered during the annotation process: 1) The descriptions of symptoms varied widely across patients, making consistent annotation difficult. For instance, while some notes explicitly mentioned concrete symptoms such as ``loss of appetite,' others used ambiguous terms like ``weight concerns'' without specifying whether the change was related to increased or decreased appetite, 2)  Pediatric patients often struggled with self-awareness and articulation of certain symptoms, particularly those requiring introspection, such as guilt or self-loathing. These symptoms were less frequently documented, as they require direct elicitation from clinicians during consultations; 3) Notes written by different clinicians showed diverse styles and depth of reporting. For example, nurse notes frequently captured nuanced behaviors and subtle symptoms that might not have been documented in more formal physician notes. Behavior-based symptoms, such as ``neglecting activities,'' were described in varying contexts, such as ``the patient quit drama club'' or ``the guardian needs to encourage the patient to do drama.'' 

In contrast to social media texts, where somatic signs are usually absent \cite{mendes-caseli-2024-identifying}, our clinical data explicitly mention patient general physical states and behaviors. This allowed us to capture a broader range of symptoms, including physical manifestations, which are often overlooked in digital text analysis.

We identified the following symptom categories from the clinical notes (all examples are paraphrased):

\begin{enumerate}

\item \textbf{Not going to school}: Interruptions in on-site schooling, e.g., \textit{``The patient hasn’t been to school this month.''} Related to BDI items 4 (\textit{``I am dissatisfied or bored with everything.''}) and 15 (\textit{``I can’t do any work at all.''}).

\item \textbf{Neglecting activities}: Interruptions in exercising or extracurricular activities, or absence of physical activities, e.g., \textit{``The patient didn’t roll into any after-school activities or clubs this year.''} Related to BDI items 4 and 15.

\item \textbf{No motivation}: Lack of motivation in completing day-to-day tasks, such as brushing teeth, e.g., \textit{``The patient has stopped following their diet plan.''} Related to BDI items 4 and 15.

\item \textbf{Feeling depressed}: Experiencing sadness or depression, e.g., \textit{``The patient has been feeling sad since their relative passed away.''} Related to BDI items 1 (\textit{``I am so sad and unhappy that I can’t stand it.''}) and 2 (\textit{``I feel the future is hopeless and that things cannot improve.''}).

\item \textbf{Feeling anxious}: Feeling anxious, overwhelmed, or stressed, e.g., \textit{``The patient is reporting elevated levels of stress caused by recent personal events.''} Related to BDI item 20 (\textit{``I am so worried about my physical problems that I cannot think of anything else.''}).

\item \textbf{Feeling down}: Experiencing low mood, e.g., \textit{``The patient’s mood has recently lowered.''} Related to BDI items 1 and 2.

\item \textbf{Irritability}: Anger or frustration directed at guardians or clinical staff, e.g., \textit{``The patient seemed very annoyed by our questions.''} Related to BDI item 11 (\textit{``I feel irritated all the time.''}).

\item \textbf{Mental health concerns}: General or unspecified mental health concerns, e.g., \textit{``Patient’s school counselor noted concerns regarding the patient’s mental health state.''} Not directly related to a BDI item, but associated with PHQ-2 (\textit{``Feeling down, depressed, or hopeless?''}).

\item \textbf{Sleep problems}: Issues with sleep, e.g., \textit{``The patient complains about having difficulty falling asleep at night and waking up in the morning.''} Related to BDI items 16 (\textit{``I wake up several hours earlier than I used to and cannot get back to sleep.'')} and 17 (\textit{``I am too tired to do anything.''}).

\item \textbf{High appetite}: Increased appetite or overeating, e.g., \textit{``The patient has gained 10+ pounds since their visit two months ago.''} Not directly related to a BDI item, but associated with PHQ-5 (\textit{``Poor appetite or overeating?''}).

\item \textbf{Low appetite}: Reduced appetite or weight loss, e.g., \textit{``Patient’s guardian reports that it is a daily struggle to ensure that the patient consumes their meals.''} Related to BDI items 18 (\textit{``I have no appetite at all anymore.''}) and 19 (\textit{``I have lost more than fifteen pounds.''}).

\item \textbf{Weight change}: Uncontrolled weight changes, e.g., \textit{``One of the patient’s concerns is weight change.''} Not directly related to a BDI item, but associated with PHQ-5 (\textit{``Poor appetite or overeating?''}).

\item \textbf{Little energy}: Persistent fatigue or tiredness without a clear cause, e.g., \textit{``Patient’s guardian reports that the patient is always complaining about feeling tired.''} Related to BDI item 17 (\textit{``I am too tired to do anything.''}).

\item \textbf{Self-loathing}: Negative self-image or self-worth issues, e.g., \textit{``The patient mentioned feeling worthless from time to time.''} Related to BDI items 7 (\textit{``I hate myself.''}) and 14 (\textit{``I believe that I look ugly.''}).

\item \textbf{Abnormal behavior}: Disoriented behavior or unusual speech patterns, e.g., \textit{``The patient looked disoriented at times, and their movements were erratic.''} Not directly related to a BDI item, but associated with PHQ-8 (\textit{``Moving or speaking so slowly that other people could have noticed? Or being so fidgety or restless that you have been moving around a lot more than usual?''}).

\item \textbf{Suicidal thoughts}: Suicidal ideation, e.g., \textit{``The patient has been referred to counseling due to suicidal tendencies.''} Related to BDI item 9 (\textit{``I would kill myself if I had the chance.''}).
\end{enumerate}

Annotations were performed at the sentence level, explicitly marking the presence or absence of depressive symptoms. This resulted in binary labels (positive or negative) for each symptom in the note. For instance, if a note contained any sentence indicating ``Irritability,'' the entire note was labeled positive for that symptom. For example, a hypothetical note containing ``The patient called regarding their Emergency Room visit yesterday. They stated that even though they refused to answer any of the clinicians’ questions last night, they have since calmed down and would like to come to the clinic again.'' was labeled positive for ``Irritability.''

Results of our annotation are summarized in Table~\ref{tab:df_stats}, which provides a detailed breakdown of symptom mentions mapped to PHQ-9 questions. It includes key symptom categories (e.g., sadness, loss of interest, or sleep problems) and indicates whether the observed changes were increased (↑), decreased (↓), or fluctuating (↑↓) compared to the patient’s norm. For instance, ``Not going to school'' was linked to a change in interest (PHQ Q1), with three notes averaging 14 words each. 

\paragraph{Observations on Annotation Patterns }

Notably, ``Feeling anxious'' was the most frequently annotated symptom, appearing in 14 notes, followed by ``Feeling depressed'' in 13 notes. In contrast, symptoms like ``Irritability,'' ``Self-loathing,'' and ``Neglecting activities'' were less frequently observed, appearing in only two notes each. 

Symptoms tied to sadness and anxiety, particularly those in Q2, were commonly documented because both patients and their guardians tended to disclose these feelings more readily. Progress notes often included complaints related to school performance, peer relationships, or family stress. Furthermore, the frequent co-occurrence of anxiety and depression increased the likelihood of these symptoms being recorded. 

Nurse notes emerged as particularly valuable sources of information, as they often captured nuanced behaviors and sensitive disclosures that might not have been documented in formal physician notes. For example, nurse notes frequently included observations about behaviors such as reluctance to engage in activities or fear of parental judgment, which were otherwise difficult to elicit. 

Conversely, symptoms requiring introspection, such as guilt or self-loathing (captured in Q6), were documented less frequently. These symptoms are inherently harder to observe, requiring proactive elicitation by clinicians. Pediatric patients, in particular, may lack the vocabulary or emotional insight to express such feelings, making these symptoms less likely to be noted unless specifically probed during consultations. Additionally, clinicians may focus more on quantifiable values, such as weight fluctuations and missed school days, rather than delving into more abstract or internalized symptoms. 

Behavior-based symptoms, such as “neglecting activities,” were described in highly variable contexts, making consistent annotation challenging. For example, some notes stated, “the patient quit drama club,” while others noted, “the guardian needs to encourage the patient to do drama” or even “the patient has not been skipping drama class very often.” This variability necessitated careful interpretation by annotators to ensure accurate symptom categorization. 

Lastly, prominent symptoms were sometimes recorded in an ambiguous manner. Terms like “weight concerns” or “mental health concerns” were frequently mentioned without specifying the exact nature of the problem. This ambiguity underscores the complexity of accurately detecting and categorizing symptoms in clinical notes. 

All these observations emphasize the challenge of detecting depression symptoms in the pediatric setting.

\begin{table}
\centering
\begin{tabular}{p{4cm}||p{1cm}|p{2cm}|p{1cm}|p{1cm}}
Question & PHQ & Semantics & \#, Notes & Av words\\\hline
Not going to school & Q1 & ↓ interest & 3 & 14 \\
Neglecting activities & Q1 & ↓ interest  & 2 & 15 \\
No motivation & Q1 & ↓ interest  & 9 & 19 \\\hline
Feeling depressed & Q2 & sadness & 13 & 14 \\
Feeling anxious & Q2 & sadness & 14 & 13 \\
Feeling down & Q2 & sadness & 3 & 13 \\
Irritability & Q2 & sadness & 2 & 12 \\
MH concerns & Q2 & sadness & 5 &  19 \\\hline
Sleep problems & Q3 & ↓↑ sleep & 7 & 16 \\\hline
High appetite & Q4 & ↑ appetite & 5 & 24 \\
Low appetite & Q4 & ↓ appetite & 7 & 16 \\
Weight change & Q4 & ↓↑ appetite & 8 & 20 \\\hline
Little energy & Q5 & ↓ energy & 3 & 14 \\\hline
Self-loathing & Q6 & ↓ self-image & 2 & 12 \\\hline
Abnormal behavior & Q8 & ↓↑ activity & 4 & 19 \\\hline
Suicidal thoughts & Q9 & suicidality & 4 & 19 \\\hline
\end{tabular}
\caption{\label{tab:df_stats} Summary of annotated positive symptom mentions mapped to PHQ-9 questions, highlighting symptom semantics (e.g., sadness, loss of interest) and whether changes were observed as increased (↑), decreased (↓), or fluctuating (↑↓) compared to the patient’s norm. For instance, ``Not going to school'' is linked to loss of interest (PHQ Q1) with 3 notes averaging 14 words each.}
\end{table}

\subsection{Natural Language Processing (NLP) Models }

Following previous research~\cite{Alsentzer2023,ziems-css}, we used the FLAN T5 (\texttt{google/flan-t5-small}) \cite{https://doi.org/10.48550/arxiv.2210.11416} model from HuggingFace in its default configuration. We also used the quantized Llama 3-70B (\texttt{neuralmagic/Meta-Llama-3-70B-Instruct-quantized.w8a16})~\cite{llama3modelcard} and Phi 3.5 mini \cite{abdin2024phi3technicalreporthighly} (\texttt{microsoft/Phi-3.5-mini-instruct}) models in their default configuration to extract symptom mentions from text. Considering the constraints of handling private clinical data and limited computational resources, we selected these models for their ability to be deployed locally, minimizing external exposure.

Due to token limitations in the Llama and Phi models, we restricted the input size to 6,000 characters, resulting in truncation for 18\% of the notes. Entire notes were fed one by one into each LLM, and we asked whether each text contained evidence of specific symptoms (binary classification). We describe the prompts we used for each model in Tables~\ref{tab:prompt1}, \ref{tab:prompt2} and \ref{tab:prompt3} below.
 
According to best practices in the domain, we used few-shot prompting for \texttt{Phi} and \texttt{Llama}, using three in-context examples. These examples were designed based on insights gained from observing the data. For \texttt{Flan} prompts, we did not use few-shot examples, again adhering to standard practices. 

As a baseline, we implemented a rigid word-level mapping approach using key parts of the questions. For instance, for the question \textit{``Does it contain evidence that the patient is not going to school?''} we matched using the words \textit{``going''} and \textit{``school.''}

\begin{table}
\begin{center}
\scalebox{0.9}{
\begin{tabular}{p{10cm}}
\textbf{System}: You are a medical AI assistant. \\
\textbf{User}: Here is an EHR note: 'Chez Dias is 15 y.o. Latina woman. Chief Complaints: SoB, Overeating, Acid Reflux.' \\
Does it contain evidence that the patient is not going to school? \\
\textbf{Assistant}: No. \\ 
\\
\hline 
\\
\textbf{User}: Here is an EHR note: 'Jeff has not been able to go out of house due to his fear of crashing during an episode. \\ 
He had to be home-schooled this year.'\\
Does it contain evidence that the patient is not going to school? \\
\textbf{Assistant}: Yes: 'He had to be home-schooled this year.' 
\\
\hline 
\\
\textbf{User}: Here is an EHR note: 'Does your child skip school? \\
Are they running around town? Do you think they are using drugs? \\
Call the Hospital to find out the best ways of taking care of your teenager or visit our website.' \\
Does it contain evidence that the patient is not going to school? \\
\textbf{Assistant}: No. 
\\
\hline 
\\
\textbf{User}: Here is an EHR note: \textbf{\emph{note text}}. Does it contain evidence that the patient is not going to school? \\
\textbf{Assistant}: \\
\hline 
\end{tabular}}
\caption{\label{tab:prompt1} Sample few-shot \texttt{Llama 3} Prompt for ``Not going to school'' (real text was paraphrased).}
\end{center}
\end{table}

\begin{table}
\scalebox{0.9}{
\begin{tabular}{p{10cm}}
\textbf{System}: You are a medical AI assistant. \\
\textbf{User}: Here is an EHR note: 'Chez Dias is 15 y.o. Latina woman. Chief Complaints: SoB, Overeating, Acid Reflux.' \\
Does it contain evidence that the patient has quit an activity or doesn't do activities? \\
\textbf{Assistant}: No. \\
\\
\hline 
\\
\textbf{User}: Here is an EHR note: 'Alice had to drop out of the knitting club this year. \\
She is anxious about the exams and fears that her family might be angry at her if she fails.' \\
Does it contain evidence that the patient has quit an activity or doesn't do activities? \\
\textbf{Assistant}: Yes: 'Alice had to drop out of the knitting club this year.'\\
\\
\hline 
\\
\textbf{User}: Here is an EHR note: 'Does your child skip school? \\
Are they running around town? Do you think they are using drugs? \\ Call the Hospital to find out the best ways of taking care of your teenager or visit our website.' \\
Does it contain evidence that the patient has quit an activity or doesn't do activities? \\
\textbf{Assistant}: No. \\
\\
\hline 
\\
\textbf{User}: Here is an EHR note: \textbf{\emph{note text}}. \\
Does it contain evidence that the patient is not going to school? \\
\textbf{Assistant}:\\
\end{tabular}}
\caption{\label{tab:prompt2} Sample few-shot \texttt{Phi} Prompt for ``Neglecting activities'' symptom (real text was paraphrased).}
\end{table}

\begin{table}
\scalebox{0.9}{
\begin{tabular}{p{10cm}}
\textbf{Premise}: This is an EHR note: \textbf{\emph{note text}}. \\
\textbf{Hypothesis}: The patient is not taking proper care of themselves and their health. \\
Does the premise entail the hypothesis? \\
\end{tabular}}
\caption{\label{tab:prompt3} Sample few-shot \texttt{Flan} Prompt for ``No motivation'' symptom (real text was paraphrased)}
\end{table}

\section{Results}

\paragraph{LLM Extraction Quality}

We assess the quality of LLM extracts by calculating F1, precision, and recall at the note level for the positive class. Precision quantifies the proportion of sentences identified as relevant by the model (positive predictions) that are actually relevant. Recall indicates the proportion of relevant sentences in the dataset that were successfully identified by the model. The F1 score is a metric that combines precision and recall into a single measure, representing the harmonic mean of the two, and is used to evaluate a model's balance between false positives and false negatives.


\begin{table}
\centering
\begin{tabular}{p{3cm}||p{1cm}|p{1cm}|p{1cm}||p{1cm}}
Category & Flan & Phi & Llama 3 70b & Baseline \\\hline
Not going to school & \bf 1.00 & 0.5 & 0.38 & \bf 1.00 \\
Neglecting activities & \bf 1.00 & 0.5 & 0.2 & 0.00 \\
No motivation & \bf 1.00 & 0.25 & 0.26 & N/A \\\hline
Feeling depressed & \bf 1.00 & 0.62 & 0.50 & 0.80 \\
Feeling anxious & \bf 0.83 & 0.77 & 0.61 & N/A \\
Feeling down & \bf 0.40 & 0.12 & 0.07 & 0.09 \\
Irritability & \bf 1.00 & 0.50 & 0.14 & N/A \\
MH concerns & 0.25 & 0.13 & 0.15 & N/A \\\hline
Sleep problems & \bf 1.00 & 0.80 & 0.33 & 0.25 \\\hline
High appetite & \bf 1.00 & N/A & 0.57 & N/A \\
Low appetite & \bf 0.71 & 0.60 & 0.54 & N/A \\
Weight change & 0.60 & \bf 0.70 & 0.24 & 0.50 \\\hline
Little energy & 0.43 & 0.29 & 0.27 & \bf 1.00 \\\hline
Self-loathing & \bf 0.67 & 0.33 & 0.33 & N/A \\\hline
Abnormal behavior & \bf 1.00 & 0.00 & 0.33 & N/A \\\hline
Suicidal thoughts & \bf 0.60 & 0.50 & 0.31 & N/A \\\hline
Average & \bf 0.78 & 0.44 & 0.33 & 0.52 \\
\end{tabular}
\caption{\label{tab:prec} Positive Precision of Symptom Mention Detection over the In-House Test Set. N/A stands for zero division.}
\end{table}

\begin{table}
\centering
\begin{tabular}{p{3cm}||p{1cm}|p{1cm}|p{1cm}||p{1cm}}
Category & Flan & Phi & Llama 3 70b & Baseline \\\hline
Not going to school & 0.33 & 0.67 & \bf 1.00 & 0.33 \\
Neglecting activities & 0.50 & \bf 1.00 & \bf 1.00 & 0.00 \\
No motivation & 0.22 & 0.67 & \bf 0.89 & 0.00 \\\hline
Feeling depressed & 0.38 & 0.77 & \bf 0.92 & 0.31 \\
Feeling anxious & 0.36 & 0.71 & \bf 1.00 & 0.00 \\
Feeling down & \bf 0.67 & \bf 0.67 & \bf 0.67 & 0.33 \\
Irritability & \bf 0.50 & \bf 0.50 & \bf 0.50 & 0.00 \\
MH concerns & 0.40 & 0.60 & \bf 1.00 & 0.00 \\\hline
Sleep problems & 0.86 & 0.57 & \bf 1.00 & 0.29 \\\hline
High appetite & 0.20 & 0.00 & \bf 0.80 & 0.00 \\
Low appetite & 0.71 & 0.86 & \bf 1.00 & 0.00 \\
Weight change & 0.38 & \bf 0.88 & \bf 0.88 & 0.25 \\\hline
Little energy & \bf 1.00 & 0.67 & \bf 1.00 & 0.67 \\\hline
Self-loathing & \bf 1.00 & 0.50 & \bf 1.00 & 0.00 \\\hline
Abnormal behavior & 0.50 & 0.00 & \bf 0.75 & 0.00 \\\hline
Suicidal thoughts & \bf 0.75 & 0.50 & 1.00 & 0.00 \\\hline
Average & 0.55 & 0.60 & \bf 0.90 & 0.14 \\
\end{tabular}
\caption{\label{tab:recall} Positive Recall of Symptom Mention Detection over the In-House Test Set. N/A stands for zero division.}
\end{table}

\begin{table}
\centering
\begin{tabular}{p{3cm}||p{1cm}|p{1cm}|p{1cm}||p{1cm}}
Category & Flan & Phi & Llama 3 70b & Baseline \\\hline
Not going to school & 0.50 & \bf 0.57 & 0.55 & 0.50 \\\hline
Neglecting activities & \bf 0.67 & \bf 0.67 & 0.33 & N/A \\\hline
No motivation & 0.36 & 0.36 & \bf 0.40 & N/A \\\hline
Feeling depressed & 0.55 & \bf 0.69 & 0.65 & 0.45 \\\hline
Feeling anxious & 0.50 & 0.74 & \bf 0.76 & N/A \\\hline
Feeling down & \bf 0.50 & 0.20 & 0.13 & 0.14 \\\hline
Irritability & \bf 0.67 & 0.5 & 0.22 & N/A \\\hline
MH concerns & \bf 0.31 & 0.21 & 0.26 & N/A \\\hline
Sleep problems & \bf 0.92 & 0.67 & 0.50 & 0.27 \\\hline
High appetite & 0.33 & N/A & \bf 0.67 & N/A \\\hline
Low appetite & \bf 0.71 & \bf 0.71 & 0.70 & N/A \\\hline
Weight change & 0.47 & \bf 0.78 & 0.38 & 0.33 \\\hline
Little energy & 0.60 & 0.40 & 0.43 & \bf 0.80 \\\hline
Self-loathing & \bf 0.80 & 0.40 & 0.50 & N/A \\\hline
Abnormal behavior & \bf 0.67 & N/A & 0.46 & N/A \\\hline
Suicidal thoughts & \bf 0.67 & 0.50 & 0.47 & N/A \\\hline
Average & \bf 0.65 & 0.51 & 0.48 & 0.22 \\\hline
\end{tabular}
\caption{\label{tab:f1} Positive F1 of Symptom Mention Detection over the In-House Test Set. N/A stands for zero division.}
\end{table}

We report results in Tables~\ref{tab:prec},~\ref{tab:recall} and~\ref{tab:f1} for the three metrics respectively.

Our overall observation suggests that all the LLMs outperform the naive word-match baseline (0.55 average F1 over LLMs vs 0.22 average F1 of our baseline). Flan achieves the highest average F1 score (0.65), followed closely by Phi (0.61). Llama 3-70B scores lowest overall (0.48). \texttt{Flan} demonstrates consistent performance across most categories, with high scores in ``Sleep problems'' (0.92) and ``Self-loathing'' (0.8), showing its strength in precision. However, it performs moderately in some categories such as ``Feeling depressed'' (0.55), ``Feeling anxious'' (0.50) and ``Weight change'' (0.47).
\texttt{Phi} performs competitively, excelling in ``Feeling depressed'' (0.69), ``Feeling anxious'' (0.74), and ``Weight change'' (0.78), indicating tendencies in balanced precision/recall.
\texttt{Llama} also shows particular strength in more general-domain categories like "Feeling anxious" (0.76) and "High appetite" (0.67), but its performance varies more significantly across other categories, such as very low scores in "Irritability" (0.22) and "Feeling down" (0.13).

Confirming our previous observations, \texttt{Flan} consistently achieves the highest average precision (0.78), outperforming \texttt{Phi} (0.44) and \texttt{Llama} 3 (0.33). This suggests \texttt{Flan} is better for the settings where false positives are not desirable due to time constraints.

\texttt{Llama 3} leads in recall with an average of 0.90, significantly higher than \texttt{Phi} (0.60) and \texttt{Flan} (0.55). Llama's high recall can be attributed to its tendency to overgeneralize when depressive symptoms are present, assuming that all symptoms are applicable across various depression-related queries. 
For example, if a general feeling of sadness is mentioned in the notes, \texttt{Llama} may incorrectly infer that the patient is overeating, due to the correlation between the two symptoms. Also, in one of the notes the mention of anxiety led \texttt{Llama} to infer irritability.
\texttt{Phi} ranks as the second-best model for both precision and recall. 

Regarding symptom-specific insights, symptoms like ``not going to school'', ``neglecting activities'', ``no motivation'', ``feeling depressed'', ``irritability'', ``sleep problems'', ``high appetite'' and ``abnormal behavior'' are detected with perfect precision by 
\texttt{Flan}, reflecting more distinct presentations of these symptoms. 
\texttt{Flan} gives the lowest precision for ``MH concerns'' (0.25) and ``little energy'' (0.43). Those categories often involve broader language, encompassing a wide range of mental health-related observations that may not directly correspond to depressive symptoms. For example, broad references to ``mental health issues in a note containing past diagnosis of Bipolar Disorder or ADHD could be discarded by the model in relation to the context.

``Not going to school'', ``neglecting activities'', ``feeling anxious'', ``MH concerns'', ``sleep problems'', ``low appetite'', ``little energy'', ``self-loathing'' and ``suicidal thoughts''. ``abnormal behavior'' show the perfect recall for \texttt{Llama 3}.  This is likely due to the model's tendency to broadly associate any general depressive language with these categories.  For example, phrases like ``struggling with everyday tasks'' or ``feeling overwhelmed'' might trigger multiple associations, leading to higher recall at the expense of precision. 

``Feeling down'' (0.67), ``irritability'' (0.50) and ``abnormal behavior'' (0.75) show the lowest \texttt{Llama 3} recall. This lower recall can be attributed to several factors. ``Feeling down'' is, on the one hand, often expressed in less direct or subtle terms compared to other symptoms.  ``Irritability'' is a symptom that can overlap with other emotional states, such as frustration or anger, and hence not picked up by the model. ``Abnormal behavior'' appears primarily in highly specific contexts, such as detailed behavioral observations by clinicians, and can be mistaken for some other external factors. 

In spite of the successes, we noticed that LLMs faced challenges in analyzing clinical notes due to their complex structure, where mixed content such as summarized complaints, clinician advice, and historical visit details can obscure relevant context. Another challenge stems from sections listing test results, particularly when only the final PHQ-9 score is presented. When the score is elevated, models tend to overgeneralize, incorrectly assuming that all depression-related symptoms apply to the patient.

In conclusion, \texttt{Flan} demonstrates good precision, making it effective for accurate and targeted symptom identification in time-sensitive screenings, while \texttt{Phi}’s higher recall makes it more suited for comprehensive screening tasks. \texttt{Llama}’s tendency to overgeneralize make it less competitive for our extraction task.

\paragraph{Validation of the Utility of LLM Extracts for Depression Screening}

Our Large Language Model (LLM) annotations aim to provide meaningful insights that could potentially aid in clinical screening. However, directly evaluating their utility with real-life doctors is challenging due to the complexity and variability of clinical workflows. As a practical proxy for human judgment, we assess the usefulness of these annotations by using them as features in a range of traditional machine learning (ML) algorithms designed for clinical decision-making. This approach allows us to evaluate whether the annotations meaningfully contribute to distinguishing between case and control patients, which is an indirect measure of their potential impact in real-world clinical settings.

For this experiment, we focused on the 15-17-year-old cohort from Table~\ref{tab:overall_phq_stats}. We extracted 3,000 clinical notes recorded within two weeks of completing the PHQ-9 questionnaire for 462 case patients. We matched by age and gender, ensuring they are born within 30 days of the corresponding case patient (462 controls with 3,000 notes). Additionally, controls had no history of a depression diagnosis up to the time of the case patient's diagnosis and had a clinical visit within the same 18-month window. This process resulted in a dataset comprising 924 patients (50/50 cases and controls)


Building on our extraction quality annotation experiment, we used \texttt{Flan} to generate a symptom vector for each patient. This vector consisted of 16 dimensions (one dimension per question). Each dimension indicated the proportion of clinical notes of a patient containing specific symptoms as detected by \texttt{Flan}. These symptom vectors served as features for a range of traditional classifiers, including Random Forest, Logistic Regression, Support Vector Machine, Multi-layer Perceptron, and Decision Tree models. The goal was to determine whether the features derived by symptom detection could be distinctive enough to separate case and control patients.

As a baseline method, we employed the Big Bird Transformers model \cite{zaheer2021bigbirdtransformerslonger} with a 512-character limit and 5-fold cross-validation to classify raw clinical note texts merged per patient without using any feature extraction method. The 5-fold cross-validation has allowed us to produce predictions over the entire dataset without explicitly allocating any data for training.

Table~\ref{tab:flan_note_num} illustrates the number of clinical notes associated with each symptom detected by \texttt{Flan}. The most frequently detected symptoms were mental health (MH) concerns (433 notes), followed by feeling down (411 notes), and sleep problems (396 notes). The overrepresentation of certain symptoms, such as sleep-related issues, may result from the mandatory documentation of these topics in clinical notes even in the absense of any particular issues. These results suggest that general mental health issues and mood-related concerns are better documented in clinical notes than we have seen in our initial annotation sample. Conversely, symptoms like high appetite (57 notes), abnormal behavior (88 notes), and no motivation (68 notes) were least frequently detected. The table highlights significant differences in the prevalence of symptoms between cases and controls, with an overall average of 72\% of symptoms documented in cases and 28\% in controls. Symptoms such as Feeling depressed (86\%), Feeling down (83\%), Self-loathing (88\%) and Suicidal thoughts (85\%) are mainly concentrated in case notes confirming mental health issues. Symptoms like Sleep problems (66\% Case, 34\% Control) and MH concerns (63\% Case, 37\% Control) show a more balanced distribution due to the documentation practices mentioned above.  

Overall, we observed very good quality of the extracts provided by Flan. For instance, the system identified ``Not going to school'' as being linked to 25 absences due to chronic illness flare-ups. Similarly, symptoms like low appetite were flagged describing urgent referrals to dietitians and psychologists: ``Patient's guardian is highly concerned with their weight and eating habits. They consulted a dietitian who urgently referred them to a psychologist. The patient denies feeling anorexic or having body image problems'' (paraphrased). Lastly, the detection of suicidal thoughts highlights provided some good insights as well: ``Patient's guardian called stating that the patient is having thoughts about self-harm. They believe that the patient is safe with them at the moment, but requires an urgent appointment.''

The observed errors in symptom detection by \texttt{Flan} often align with interpretations of due to figurative language. For example, when a note mentioned that a patient had a ``good appetite'' as part of their recovery from an illness, \texttt{Flan} incorrectly detected overeating. 

\begin{table}
\centering
\begin{tabular}{p{3.5cm}||p{1.5cm}|p{1.5cm}|p{1.5cm}}
Question & \#Notes All & \%, Case & \%, Control \\\hline
Not going to school & 91 & 59 & \bf 41 \\ 
Neglecting activities & 210 & 60 & 40 \\ 
No motivation & 68 & 79 & 21 \\\hline  
Feeling depressed & 280 & 86 & 14 \\ 
Feeling anxious & 271 & 78 & 22 \\ 
Feeling down & \bf 411 & 83 & 17 \\ 
Irritability & 109 & 67 & 33 \\ 
MH concerns & \bf 433 & 63 & 37 \\\hline  
Sleep problems & 396 & 66 & 34 \\\hline  
High appetite & 57 & 74 & 26 \\ 
Low appetite & 291 & 68 & 32 \\ 
Weight change & 107 & 60 & 40 \\\hline  
Little energy & 271 & 63 & 37 \\\hline 
Self-loathing & 234 & \bf 88 & 12 \\\hline  
Abnormal behavior & 88 & 65 & 35 \\\hline  
Suicidal thoughts & 279 & 85 & 15 \\\hline\hline
All positive notes / Average &  3596 & 72 & 28 \\\hline
\end{tabular}
\caption{\label{tab:flan_note_num} Number of notes per symptom detected by \texttt{Flan} for differential diagnosis.}
\end{table}

\begin{table}
\centering
\begin{tabular}{p{5.5cm}||p{2cm}|p{1.2cm}|p{1.2cm}|p{1.2cm}}
Model & AUC-ROC & F1 Score & Precision & Recall \\  \hline
Random Forest (RF) & 0.69 & 0.62 & 0.72 & 0.54 \\
Logistic Regression (LR) & 0.68 & 0.58 & \bf 0.78 & 0.47 \\ 
Support Vector Machine (SVM) & \bf 0.71 & \bf 0.63 & \bf 0.78 & 0.53 \\ 
Multi-layer Perceptron (MLP) & 0.70 & 0.62 & 0.74 & 0.53 \\ 
Decision Tree (DT) & 0.63 & 0.59 & 0.71 & 0.50 \\\hline \hline
BERT full text baseline & 0.60 & 0.45 & 0.43 & \bf 0.55 \\
\end{tabular}
\caption{\label{tab:class_res} Case-control classification results: Differential Diagnosis use case.}
\end{table}

Table~\ref{tab:class_res} presents the classification performance of a range of ML models in distinguishing between case and control groups. Among the models tested, Support Vector Machine (SVM) achieved the highest overall performance, with an AUC-ROC of 0.71, an F1 score of 0.63, and a precision of 0.78. The average AUC-ROC across five models is 68, which is 8\% increase as compared to the baseline performance (AUC-ROC 0.60). These findings confirm the utility and discriminative power of our LLM-based feature extraction procedure.

\section{Discussion}

\paragraph{Statement of the key findings}

This study explored the potential of using Large Language Models (LLMs) to identify depressive symptoms within pediatric electronic health records (EHRs). By benchmarking the zero-shot performance of three state-of-the-art LLMs—FLAN-T5, Phi, and Llama 3-70B—on extracting PHQ-9-related symptoms from free-form text, we demonstrate the potential of these models to improve the consistency of depression screening in the challenging pediatric context. 

Our key findings showed that  LLMs, particularly FLAN-T5  achieved high precision (up to 80\% across 16 symptom categories) in identifying depressive symptoms. These results suggest that LLMs can support automated mental health screening by providing accurate symptom extraction, which can be useful for automated mental health screening.

\paragraph{How this study adds / supports previous studies in this area?}

This study builds on prior research by applying large language models (LLMs) to annotate depressive symptoms directly from clinical notes~\cite{chim-etal-2024-overview,wang-etal-2024-explainable,Lan2024, Rosenman2024}.

This study supports the utility of zero-shot LLMs like FLAN-T5 for extracting symptom-related evidence~\cite{Alsentzer2023,ziems-css}, a method previously underexplored in paediatric mental health contexts. 

The efficacy and safety of such tools in real-world clinical settings remain to be thoroughly investigated~\cite{Omar2025}. This study highlights the advantage of utilizing real-world data to assess these tools, particularly when applied to depression. While majority of studies suggest that LLMs cannot replace human therapists~\cite{Minerva2023-ll,Khawaja2023-tt}, they can serve as supplementary tools. These tools can integrate human clinical insights to improve therapeutic processes.

However, challenges remain, including issues related to data bias and the necessity of human oversight. These challenges emphasize the need for careful integration of LLMs into existing healthcare frameworks~\cite{Levkovich2023-cq,Heston2023-zk}. It is crucial to augment rather than replace traditional diagnostic and treatment practices to ensure the effective and safe use of these tools in clinical settings, especially for treating depression. And we have proposed an LLM-based approach to detect PHQ-9 symptoms from unstructured data, such as doctor observations and patient replies, to support traditional practices.

\paragraph{Strengths and Limitations}

The strengths of this work lie in its novel application of LLMs for text-based evidence extraction in a domain where annotated data is particularly sparse, and texts are very heterogeneous (mental health clinical notes). We also use the LLM extracts as features to train an automated screening system as a proxy to human screening, which outperforms a strong baseline not using our feature extracts.

By leveraging the generalization capabilities of LLMs in a zero-shot setting, this approach avoids the need for costly and time-consuming manual annotation of training datasets. Furthermore, the ethical framework employed ensures that the models are used solely for evidence extraction, maintaining clinician agency and control over the diagnostic process. Also, the best-precision model FLAN-T5, being a relatively small and computationally efficient model, is well-suited for deployment in clinical settings where resources are limited.

The study has some limitations.   First, while the zero-shot framework eliminates the need for annotated training data, it may lead to variations in performance across different clinical settings or note structures. This requires further validation. Second, the study focuses on extracting PHQ-9-relevant information, which, while validated, does not encompass the full spectrum of depressive symptoms or contextual factors. Additionally, our results are constrained to the pediatric age group (6–24 years), and further research is needed to generalize these findings to broader populations or other mental health conditions.

\paragraph{Implications}

The implications of this study are significant for the field of mental health diagnostics, particularly in pediatric care. By addressing the challenges of data scarcity and heterogeneity, LLM-based tools can enhance the efficiency and consistency of depression screening, particularly in resource-limited settings. Moreover, the interpretability of extracted text segments allows for greater transparency and facilitates clinicians' trust in AI-assisted tools. This approach also provides a foundation for future research into using LLMs for feature extraction for downstream AI across mental health conditions.

\subsection{Conclusion}

In conclusion, this work highlights the potential of LLMs to serve as ethical, efficient, and scalable tools for evidence extraction in pediatric mental health diagnostics. While challenges remain in ensuring reliability and generalizability, the findings underscore the promise of integrating LLMs into clinical workflows to enhance the detection and management of depression in young patients. Further research should focus on refining these models, addressing their limitations, and validating their performance in real-world clinical settings.

\backmatter

\section*{Code and Data Availability}

The code and data could not be publicly shared due to their confidentiality requirement. The code was implemented in Python 3.10.

\section*{Acknowledgments}

This work was funded by Cincinnati Children's Hospital Medical Center's Mental Health Trajectory program. The views expressed are those of the authors and not necessarily those of the Cincinnati Children's Hospital Medical Center's Decode program. This work was authored in part by UT-Battelle, LLC, under Contract No. DE-AC05-00OR22725 with the U.S. Department of Energy. The United States Government retains and the publisher, by accepting the article for publication, acknowledges that the United States Government retains a non-exclusive, paid-up, irrevocable, world-wide license to publish or reproduce the published form of this manuscript or allow others to do so, for United States Government purposes. The Department of Energy will provide public access to these results of federally sponsored research in accordance with the DOE Public Access Plan (http://energy.gov/downloads/doe-public-access-plan). 

\section*{Author contributions statement}

MI, JI and JP conceived the approach and experiments. DS prepared the datasets. MI processed the datasets, conducted the experiments, analyzed the results, and wrote the paper. PB assisted in writing the manuscript. All reviewed the research and the manuscript. All authors approved the manuscript.

\section*{Additional information}

\textbf{Competing interests}: The work is in collaboration with Cincinnati Children's Hospital Medical Center, University College London and Queen Mary University of London.

\begin{appendices}




\end{appendices}


\bibliography{sn-bibliography}

\end{document}